%% file: main_arxiv.tex
\documentclass{article}

\usepackage{microtype}
\usepackage{graphicx}
\usepackage{subfigure}
\usepackage{booktabs} %

\usepackage{hyperref}

\usepackage[arxiv]{icml2024}

\usepackage{amsmath}
\usepackage{amssymb}
\usepackage{mathtools}
\usepackage{amsthm}

\usepackage[capitalize,noabbrev]{cleveref}

\usepackage{listings}
\usepackage{url}
\usepackage{comment}
\usepackage{xspace}

\usepackage{algorithm}

\theoremstyle{plain}

\theoremstyle{definition}

\theoremstyle{remark}

\usepackage[textsize=tiny]{todonotes}
\usepackage[htt]{hyphenat}

\newcommand{\TASKNAME}{UFO\xspace}
\newcommand{\METHODNAME}{L3GO\xspace}
\newcommand{\ENVNAME}{SimpleBlenv\xspace}

\definecolor{ao}{rgb}{0.0, 0.5, 0.0}

\icmltitlerunning{L3GO: Language Agents with Chain-of-3D-Thoughts for Generating Unconventional Objects}

\begin{document}

\twocolumn[

\icmltitle{
  L3GO: Language Agents with Chain-of-3D-Thoughts for Generating Unconventional Objects
}

\icmlsetsymbol{equal}{*}

\begin{icmlauthorlist}
\icmlauthor{Yutaro Yamada}{yyy}
\icmlauthor{Khyathi Chandu}{comp}
\icmlauthor{Yuchen Lin}{comp}
\icmlauthor{Jack Hessel}{comp}
\icmlauthor{Ilker Yildirim}{yyy}
\icmlauthor{Yejin Choi}{comp,sch}
\end{icmlauthorlist}

\icmlaffiliation{yyy}{Yale University}
\icmlaffiliation{comp}{Allen Institute for AI}
\icmlaffiliation{sch}{University of Washington}

\icmlcorrespondingauthor{Yutaro Yamada}{yutaro.yamada@yale.edu}

\icmlkeywords{Machine Learning, ICML}

\vskip 0.3in
]

\printAffiliationsAndNotice{}  %

\input{sec/0_abstract}
\input{sec/1_intro}

\input{sec/2_related}

\input{sec/4_method}

\input{sec/5_experiment_shapenet}

\input{sec/6_experiment_unusual}
\input{sec/8_ablation}
\input{sec/11_conclusion}

\input{sec/impact_statement}
\bibliography{zotero,example_paper}
\bibliographystyle{icml2024}

\newpage
\appendix

\onecolumn

\input{sec/12_appendix}

\input{sec/9_app_future}
\input{sec/10_limitation}

\end{document}

%% file: sec/0_abstract.tex
\begin{abstract}

Diffusion-based image generation models such as DALL-E 3 and Stable Diffusion-XL demonstrate remarkable capabilities in generating images with realistic and unique compositions. %
Yet, these models are not robust in precisely reasoning about physical and spatial configurations of objects, especially when instructed with unconventional, thereby out-of-distribution descriptions, such as \emph{``a chair with five legs''}. 
In this paper, we propose a language agent with chain-of-3D-thoughts (\METHODNAME), an inference-time approach that can reason about part-based 3D mesh generation of unconventional objects that current data-driven diffusion models struggle with. More concretely, we use large language models as agents 
to compose a desired object via trial-and-error within the 3D simulation environment. To facilitate our investigation, we develop a new benchmark, \textbf{Unconventionally Feasible Objects (UFO)}, as well as \ENVNAME, a wrapper environment built on top of Blender\footnote{\url{https://www.blender.org/}} where language agents can build and compose atomic building blocks via API calls. 
Human and automatic GPT-4V evaluations show that our approach surpasses the standard GPT-4 and other language agents (e.g., ReAct and Reflexion) for 3D mesh generation on ShapeNet. 
Moreover, when tested on our UFO benchmark, our approach outperforms other state-of-the-art text-to-2D image and text-to-3D models based on human evaluation.

\end{abstract}

%% file: sec/1_intro.tex
\begin{figure}[h]
    \centering
    \includegraphics[width=\linewidth]{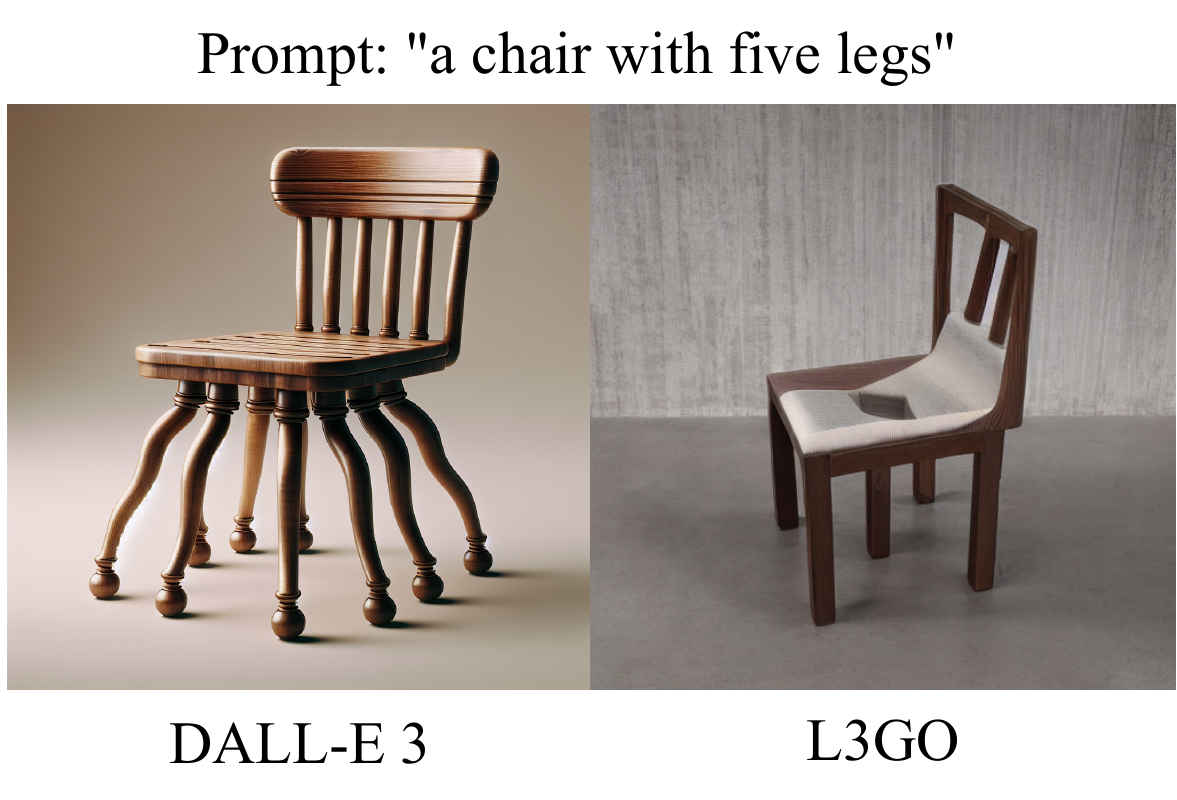}
    \vspace{-7mm}
    \caption{We compare one of the state-of-the-art text-to-image models (DALL-E 3) with our LLM-based approach (\METHODNAME). We perform five iterations of DALL-E 3 generation with human feedback but DALL-E 3 does not strictly follow the prompt. \METHODNAME creates a chair with the correct number of legs.}
    \label{fig:dall_e_3_vs_l3go}
    \vspace{-4mm}
\end{figure}

\section{Introduction}
\label{sec:intro}

AI applications that generate 2D images \cite{betkerh, saharia2022, podell2023} and 3D models \cite{jun2023, lin2023} %
from text instructions have opened up significant %
possibilities for creators.
However, these tools lack precise output controls as they often produce unexpected or ``\textit{hallucinatory}'' results \cite{saharia2022} not loyal to the input prompts.
Additionally, early versions of Stable Diffusion \cite{rombach2022} had difficulty in combining multiple concepts in one image or would mix up different attributes. %
Previous efforts have improved performance on object-attribute attachment, missing objects, etc. by steering attention layers \cite{feng2022a, chefer2023atg, rassin2023}, or, by
training larger models with detailed captions on a vast scale (StableDiffusion-XL (SDXL) \cite{podell2023} and DALL-E-3 \cite{betkerh}.) %
However, even the most performant diffusion model, DALL-E 3, still fails to generate objects that require precise 3D spatial understanding like ``a chair with five legs'' (Figure ~\ref{fig:dall_e_3_vs_l3go}).
This difficulty persists even after repeated attempts to adjust DALL-E-3’s outputs with human feedback directly, e.g., ``The chair you generated has seven legs. Please make a chair with exactly five legs.''

\begin{figure*}[!ht]
    \centering
    \includegraphics[width=\textwidth]{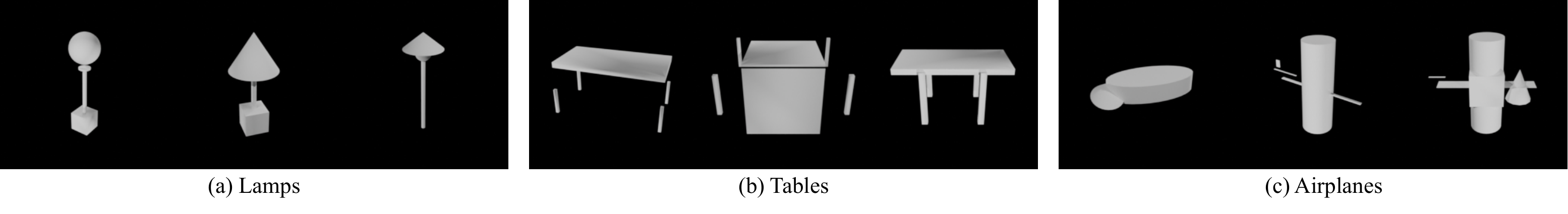}
    \vspace{-7mm}
    \caption{GPT-4 tries to construct three types of objects from ShapeNet by writing Python scripts in Blender. It can successfully create simple items like lamps, but faces challenges with more complex objects such as tables and airplanes.}
    \label{fig:gpt-4_shapenet_examples}
    \vspace{-4mm}
\end{figure*}

We posit that the sophisticated text-based reasoning abilities inherent in LLMs can compensate for shortcomings in the 3D spatial comprehension of text-to-2D image and text-to-3D models. We present \METHODNAME, an inference agent capable of iteratively soliciting feedback from LLMs to integrate corrections and enhance the precision of rendering a 3D mesh used as a skeleton to generate 2D image.

We conduct our experiments within Blender —a widely acclaimed 3D modeling software. We create and release an environment called \ENVNAME, based on Blender, to systematically evaluate text-to-3D mesh generation performance of LLM agents. State of the art LLMs such as GPT-4  \cite{bubeck2023}, despite being trained on only text has decent spatial reasoning capabilities. Figure \ref{fig:gpt-4_shapenet_examples} shows mixed results when GPT-4 is prompted to write a Python script that can run in Blender to create 3D meshes of basic objects solely based on the object name. On the one hand, text-only GPT-4 demonstrates surprising proficiency in creating simple 3D objects like lamps (2a) comprising of three basic shapes. However, as object complexity increases to more than four parts (ex: four legs of a table) or complex objects like airplane (2b, 2c), GPT4's success in perfectly assembling them is limited.

Our \METHODNAME agent bridges these gaps by breaking down the constructing complex objects by employing a more structured, part-by-part approach into: (1) identifying relevant parts specifications and critiquing them, (2) identifying spatial specifications and placement, (3) running the current action to critique the spatial placement and completion. This setup iteratively seeks feedback from \ENVNAME and the specifications and critiques are generated from LLM. Finally, we render the mesh into an image, and feed it into ControlNet \cite{zhang2023} with Canny edge detection \cite{canny1986itpami} to generate a  textured and more natural looking image. We conduct human evaluations to compare the performance of LLM-based mesh creation using 13 popular object categories from ShapeNet.
\METHODNAME outperforms basic GPT-4, ReAct-B, and Relfexion-B according to both human and auto evaluation.
We also show that mesh quality evaluation using GPT-4V \cite{openai2023} yields a metric with high correlation to human judgement. Finally, we introduce Unconventionally Feasible Objects, named \TASKNAME with unconventional yet feasible  objects. We show that \METHODNAME surpasses current state-of-the-art text-to-2D image and text-to-3D mesh models on \TASKNAME. 
Collectively, our findings indicate the promising role of integrating language agents in diffusion model pipelines, particularly for constructing objects with specific attribute requirements in the future applications of generative AI.

%% file: sec/2_related.tex
\section{Related Work}
\label{sec:related}

\paragraph{Spatial Understanding of Language Models}

Numerous studies have delved into the spatial comprehension capabilities of language models. \citet{janner2018tacl} explored the spatial reasoning of LSTM \cite{hochreiter1997nc} in a simulated environment with agent actions and rewards, though their 2D grid world environment is notably simpler than the 3D modeling context considered in our work. \citet{abdou2021p2ccnll} and \citet{patel2022} demonstrated that language models develop internal representations for directional information. Additionally, \cite{mirzaee2021p2cnacaclhlta} introduced a question-answering benchmark for spatial reasoning based on language description. It is reasonable to assert that LLMs exhibit spatial reasoning capabilities and can be effectively prompted to offer feedback on spatial constructions.

\paragraph{Large Language Models as Agents}
LLM agents \cite{ge2023, park2023p3aasuist, shen2023, gupta2023, wu2023, Yin2023LumosLA} represent a new category of artificial intelligence systems built upon large models. 
These agents are capable of acting, reasoning, and interacting with external environments. 
Although LLMs has limited executable skills on their own, when integrated with external APIs and knowledge sources, they can tackle a wide range of tasks \cite{schick2023}. %
An iterative approach has shown to be beneficial to solve natural language processing tasks, as evidenced by ReAct \cite{yao2022} and embodied tasks applied to games \cite{wang2023a, lin2023swiftsage}, web navigation \cite{yao2022anips}, and robot navigation \cite{huang2022}.
Our approach, which utilizes LLMs for creating 3D meshes, contributes to the expanding research in this developing field.

\paragraph{Text to 3D models}

A growing number of studies are exploring how to adapt pre-trained text-to-2D image diffusion models for text-to-3D generation \cite{poole2022, lin2023, wang2023}, suggesting that similar challenges are found in text-to-3D models. %
Meanwhile, LLM based approaches introduce a new perspective to text-to-3D mesh creation, potentially offering ways to address issues with out-of-distribution samples.

%% file: sec/4_method.tex
\section{\METHODNAME framework}
\label{sec:method}

\begin{figure*}[!t]
    \centering
    \vspace{-2mm}
    \includegraphics[width=\textwidth, trim={2cm 5cm 0 5.5cm},clip]%
    {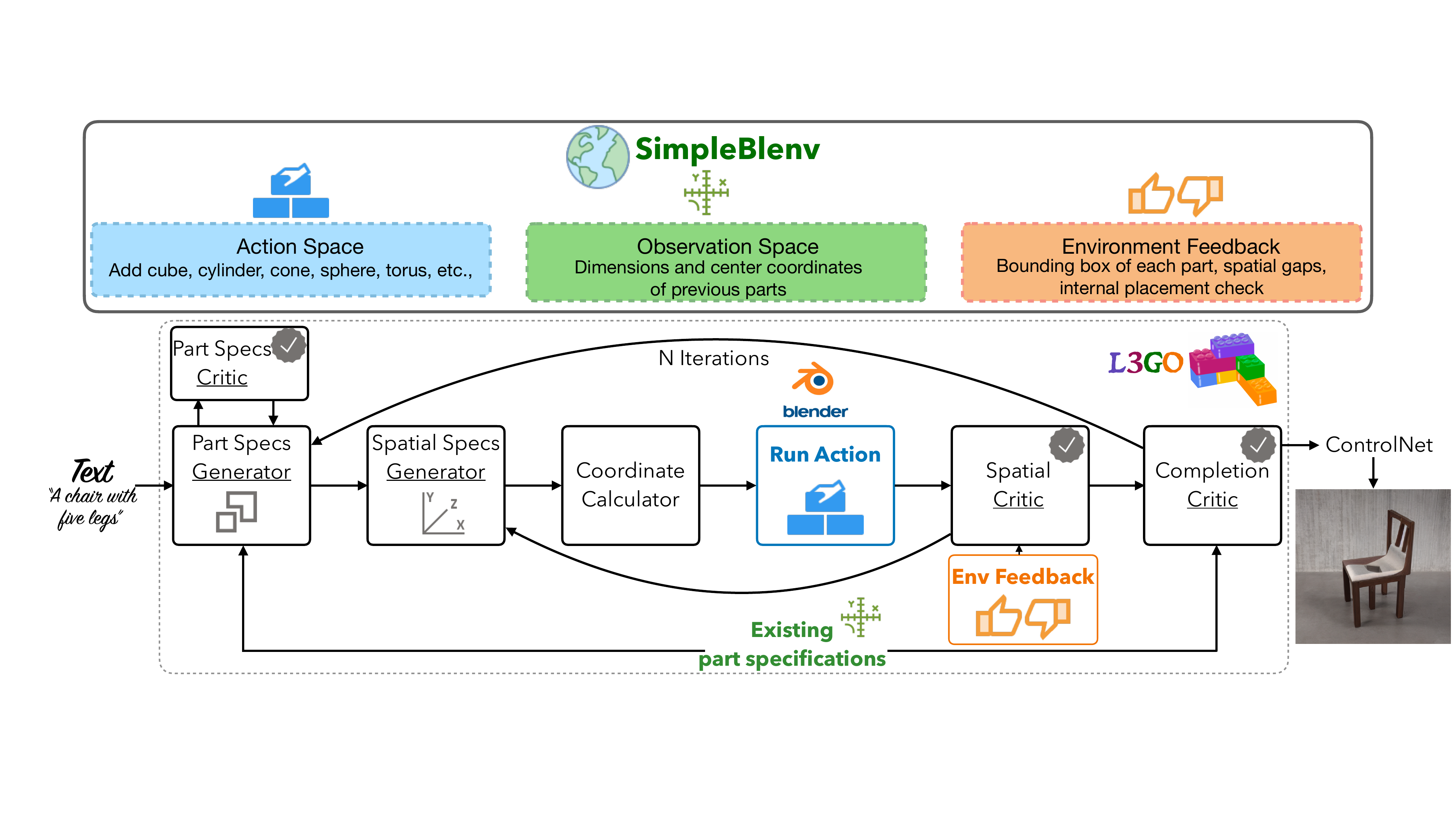}
    \vspace{-10mm}
    \caption{(Top): \ENVNAME, a wrapper environment on top of Blender, where LLM can construct a 3D mesh by using atomic building blocks. (Bottom): Schematic diagram of \METHODNAME. %
    }
    \vspace{-4mm}
    \label{fig:main}
\end{figure*}

The main challenge with generating entire 3D objects in one go is that it often leads to compounding spatial inaccuracies. 
We propose decomposing the creation of a 3D mesh into distinct parts and placing each component step by step. 
We name this approach \METHODNAME, an agent that can collect the feedback and execute the action from a chain of 3D thoughts in a simple 3D environment.
This approach transforms a singular attempt at object construction into iterative feedback collection and correction processes, enabling the integration of feedback from the Blender environment. 
Our framework borrows ideas from previous work of LLM-based reasoning  and acting \cite{yao2022anips, wang2023a}, but has been adopted for 3D mesh construction in a practical 3D modeling software. 

\subsection{\ENVNAME}
We introduce \ENVNAME, an environment built on top of Blender, where agents can easily submit action commands and receive environmental feedback, such as bounding box information and placement errors. We plan to release the code for the environment.

\paragraph{Action space:}
In Blender, professional 3D designers have the ability to create complex 3D models, with a nearly limitless range of actions at their disposal. 
Although nearly every action a user can perform in Blender's UI has a corresponding Python API, we choose to focus on five basic shape primitive APIs for the sake of simplicity. %
These APIs are:
\texttt{primitive\_cube\_add, primitive\_cylinder\_add, primitive\_cone\_add}, \texttt{primitive\_uv\_sphere\_add}, and \texttt{primitive\_torus\_add}. 
As their names imply, these are used for adding cubes, cylinders, cones, spheres, and toruses, respectively.

These API functions come with a complex set of arguments. 
To make it easier for LLMs to use these functions, we wrap each function with a wrapper that only requires a few key parameters, such as scale, location, and radius. 
Consequently, the range of actions available to our \METHODNAME agent includes different settings for size, position, radius, and so on, for each shape-creating command. 
An illustration of this can be seen in the following example:

\begin{lstlisting}[breaklines=True, frame=single]
def create_cube(name, location, scale):
    bpy.ops.mesh.primitive_cube_add(size=1, location=location)
    cube = bpy.context.object
    cube.name = name
    cube.scale = scale
    return cube
\end{lstlisting}

For a detailed list of all the action wrapper APIs we created, refer to the Appendix.
Despite using only five basic shape APIs, agents can create objects in many ways thanks to the flexibility in scaling, positioning, and adjusting the radius, among other controls.

\paragraph{Observations and Environment Feedback}

We maintain a state space representation as a list of object parts that have been created so far, including their size in terms of x, y, z-axis and location in the global coordinate.
Regarding environment feedback, after the agent selects an action, we execute the action in Blender thereby creating a mesh in the virtual space. 
From this, we can (a) extract information such as the bounding box dimensions of the object parts, and (b) check if the part built by the agent is intersecting with any other parts or if there is an unnecessary gap between the parts (e.g. see Figure \ref{fig:spatial_errors}.) 
We have built a set of functions to directly gather this information from Blender. 
This feedback is then relayed to the \METHODNAME agent as text messages before it takes its next action. %

\subsection{\METHODNAME: LLM-based 3D Generation of Objects}

In this section, we introduce \METHODNAME, an LLM agent specifically designed for 3D mesh creation from text.
\METHODNAME is comprised of six components, each powered by a language model that either functions as a \textit{generator} or a \textit{critic}. The schematic diagram in Figure \ref{fig:main} is shown for a visual overview.

\paragraph{Part Specifications Generator:}
\METHODNAME first prompts the LLM to identify the most pivotal part of the object. 
This pivotal part makes it easier to attach subsequent components. 
For instance, starting with the seat of a chair is practical because it is straightforward to add legs and a backrest to it, simplifying the coordinate calculations for the other parts. 
After naming the part, the agent uses a size generator to determine its reasonable dimensions in terms of width, depth, and height, corresponding to the x, y, and z axes.

\paragraph{Part Specifications Critic:}

Once a part name is proposed, it undergoes a review by the Part Specifications Critic. 
This step is crucial to avoid ambiguity, which can confuse the agent later. 
For example, if ``leg'' is proposed while creating a chair, the agent cannot know its exact placement without a spatial descriptor like ``front right leg''. 
The Part Specifications Critic's role is to identify and correct such vague descriptions, allowing the Part Specifications Generator to revise its suggestion accordingly. 
The process moves forward only after the Part Specifications Critic's  approval.

\paragraph{Spatial Specifications Generator:}

After establishing the part name and size, the model considers the spatial requirements of the part, given what has already been constructed. (For the first part, we simply position it at the center.) 
The agent begins by selecting the most appropriate base part to attach the new component to, then determine the spatial relationship between them. 
For instance, if constructing an airplane with the fuselage as the base and the left wing as the new part, a typical spatial requirement would be to attach the wing to the middle of the fuselage's left side.

\paragraph{Coordinate Calculator:}

Based on the spatial requirements and the base part's position, this component calculates the new part's center coordinate. 
Accuracy here is crucial, as even minor misalignments can impact the overall correctness.
To ensure precision, the agent is given access to a python execution environment: while using the LLM only to generate Python code for calculating the position of the new part. 
This approach is similar to the one described in \citet{gao2023p4icml}. 
To increase reliability, the process is repeated three times and determine the x, y, and z coordinates based on a majority vote from these repetitions (with ties broken arbitrarily)

\paragraph{Run action:}

After determining the size and spatial position, the agent asks an LLM to decide on the part's shape, choosing from cubes, cylinders, cones, spheres, and toruses. Then, the agent writes a valid Python script for Blender, specifying the size, position, and shape type. Finally, the agent runs a command to generate the part's mesh in Blender.
This code is executed in Blender in a headless mode, and the environment provides important feedback, such as the bounding boxes of each generated part, which is used in the next module.

\paragraph{Spatial Critic:}

\begin{figure}[!t]
    \centering
    \includegraphics[width=0.8\linewidth]{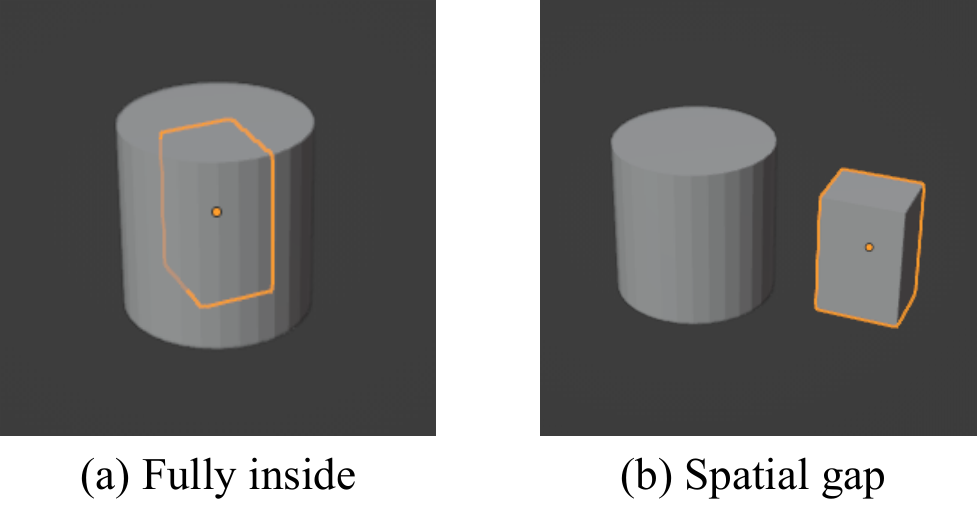}
    \vspace{-2mm}
    \caption{Two types of error feedback we provide in \ENVNAME: (a) The newly added cuboid (in orange) is completely inside the base cylinder. (b) There is unnecessary spatial gap between the newly added cuboid and the base cylinder.}
    \label{fig:spatial_errors}
    \vspace{-4mm}
\end{figure}

After running the Blender code, two final spatial correctness checks are conducted:
a \emph{continuity check}, which tells the agent if the newly created part is disconnected from the existing parts, and a \emph{total overlap check} with existing parts, which tells the agent if a newly created part is entirely contained within an existing part.
If either issue arises, the process returns to the spatial requirement generation stage and the agent adjusts accordingly.
See examples of spatial errors in Figure \ref{fig:spatial_errors}.

\paragraph{Completion Critic:}

The final step is to determine whether the construction of a 3D mesh for an object is completed. 
To do this, this critic is provided with the name of the object being built and the list of its parts that have already been constructed to an LLM to make a binary decision of completion. 
If the critic predicts that it is incomplete, we start the next iteration with the Part Specifications Generator. 
If the task is completed, we proceed to generating a more natural-looking image using ControlNet.

\paragraph{ControlNet for 3D Meshes $\rightarrow$ 2D Images}

After the \METHODNAME agent finishes the creation of a 3D mesh, we render the object into gray-scaled image.
We then feed this image to ControlNet with Canny edge detection to produce a more realistic looking image.

Note that as \METHODNAME is text-based, it does not use visual information. 
Therefore, all communication must be text-based, including defining spatial orientations like which direction is the front or back when asking the model to create an object. 
We set these spatial assumptions in the prompts in advance to guide the construction process.
Unless specified otherwise, we use GPT-4 as our base LLM in \METHODNAME in our experiments.

%% file: sec/5_experiment_shapenet.tex
\section{Experiments on ShapeNet}
\label{sec:experiment}

In this section, we detail the baseline methods (\S \ref{subsec:baseline}) to compare text-to-2D image and text-to-3D mesh generation to compare with our \METHODNAME.
We demonstrate the effectiveness of LLM-based methods in the generation of simple 3D objects, specifically focused on 13 well-known categories sourced from ShapeNet.
For automatic evaluation, we use GPT-4V as evaluator for recognizing 3D meshes (\S \ref{subsec:gpt-4v-eval}) of simple objects. 
We also show that human assessments and GPT-4V's automated evaluations (\S \ref{subsec:human_correlation}) are well correlated.

\subsection{Baselines}
\label{subsec:baseline}

Given the absence of pre-existing LLM agents designed to work in Blender, we chose a range of algorithms that serve as baseline references.
Originally intended for natural language processing tasks, we have adapted these baselines to function within the Blender environment, ensuring they align with our experimental framework.

\paragraph{ReAct-B}
ReAct \cite{yao2022} is a framework designed for implementing a language model based agent. 
In this framework, the agent outputs its thought process before taking any action. 
The observations gathered from the external environment following this action inform the subsequent steps. 
We implement ReAct in the \ENVNAME setting, utilizing the environment feedback, observations, and action space outlined in Section \ref{sec:method}. 
To differentiate it from the text version of ReAct, we refer to our implementation as ReAct-Blender, or ReAct-B for short.

\paragraph{Reflexion-B}
Reflexion \cite{shinn2023} builds upon the ReAct framework by adding an extra step of reflection. 
In ReAct, the agent outlines its reasoning, takes an action, and then receives feedback from the environment. 
Reflexion goes a step further -- after receiving environment feedback, the agent engages in reflection on the action taken and its results, to inform its next move. 
In our setup, at the end of every iteration, we consider the current object part, previously built parts, and the current environment feedback. 
We then ask the agent to reflect on the size and placement of the object part it has just built. 
After the agent shares its thoughts, we prompt it to decide whether to redo the current part or move on. 
If the agent chooses to redo, its reflective insights are used in the next step.

\paragraph{GPT-4}
For a less structured approach, we use GPT-4 to generate the entire Python code needed to create an object in one single attempt.
The prompt we used for GPT-4 was adapted from a Github repository \cite{gd3kr2023}, which, to our knowledge, was the first to present the open-source project that uses GPT-4 to control Blender. 
For the full prompt, refer to the Appendix.

\begin{table}
  \centering
  \resizebox{\linewidth}{!}{
  \begin{tabular}{@{}cccccc@{}}
    \toprule
      & Human & \METHODNAME & ReAct-B & Reflexion-B &  GPT-4 \\
    \midrule
GPT-4V & 0.877 & \textbf{0.6} & 0.423 & 0.4 & 0.346 \\
Human & 0.894 & \textbf{0.584} & 0.385 & 0.403 & 0.445 \\
    \bottomrule
  \end{tabular}
  }
  \vspace{-2mm}
  \caption{Mean accuracy of different LLM-based agents on ShapeNet-13, evaluated by GPT-4V (top row) and humans (bottom row); each cell is an average over 130 trials. `Human' in the column names refers to the original ShapeNet meshes, designed by humans, which can be considered as the upper bound. 
  We see that \METHODNAME outperforms other GPT-4-based agents (e.g. ReAct-B, Reflexion-B, and unmodified GPT-4).
  }
  \label{tab:gpt-4v_eval_shapenet-13}
  \vspace{-4mm}
\end{table}

\paragraph{Dataset: ShapeNet-13}
To assess the basic mesh generation ability, we use 13 categories of ShapeNet: [`airplane', `bench', `cabinet', `car', `chair', `display', `lamp', `loudspeaker', `rifle', `sofa', `table', `telephone', `watercraft'], as introduced by \cite{choy2016}.

\subsection{Automatic Evaluation via GPT-4V}
\label{subsec:gpt-4v-eval}

To streamline our evaluation process, we propose using GPT-4V to assess the performance of mesh construction. 
For each object category, we generate 10 meshes from GPT-4, ReAct-B, Reflexion-B, and \METHODNAME.
After the agent finishes mesh construction, we render the object from 10 different views by rotating the camera around the object at the same height.
This results in 10 images per mesh. 
we then feed 10 images to GPT-4V all at once, and use the following prompt: `\textit{`What object do you see in these images? Answer with a single object name. Your answer must be one of the following options: [airplane, bench, cabinet, car, chair, display, lamp, loudspeaker, rifle, sofa, table, telephone, watercraft]''}. 
 
\begin{figure}[!t]
    \centering
    \includegraphics[width=\linewidth]{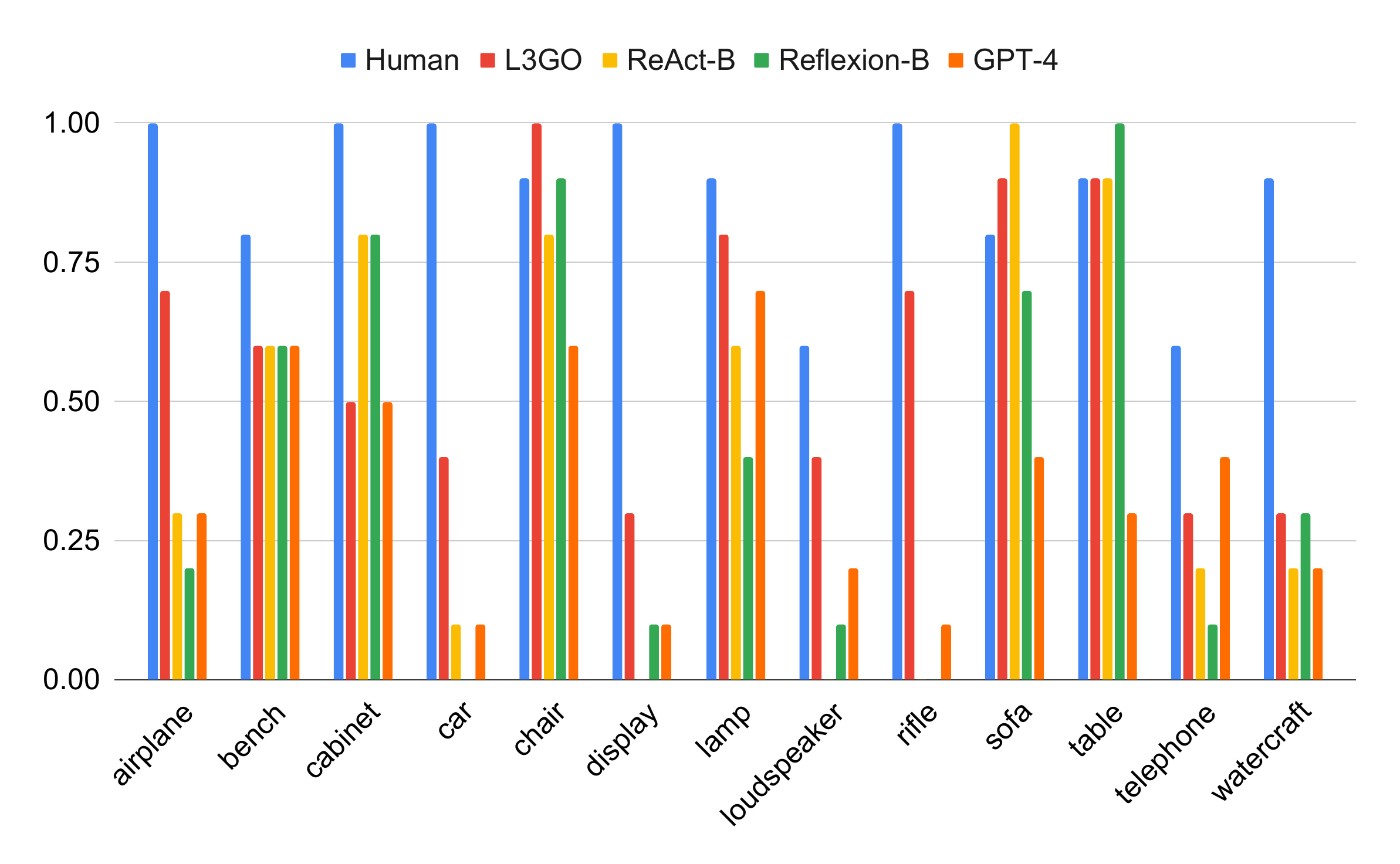}
    \vspace{-7mm}
    \caption{GPT-4V evaluation of \METHODNAME, ReAct-B, Reflexion-B, and GPT-4 on ShapeNet-13. `Human' refers to original ShapeNet meshes that were designed by humans. For complex objects such as airplanes and rifles, \METHODNAME performs better than others.}
    \label{fig:gpt-4v_eval_shapenet-13_individual}
\end{figure}

Table \ref{tab:gpt-4v_eval_shapenet-13} presents the average accuracy across different object categories. 
It is evident that structured methods, including ReAct-B, Reflexion-B, and \METHODNAME, surpass GPT-4 in performance. 
Notably, among these structured approaches, \METHODNAME proves to be the most effective. 
Delving into the results for each object, as detailed in Figure \ref{fig:gpt-4v_eval_shapenet-13_individual}, it becomes clear that \METHODNAME particularly excels in constructing complex objects like airplanes and rifles.

\paragraph{Correlation with human evaluations}
\label{subsec:human_correlation}

\begin{figure}[!t]
    \centering
    \includegraphics[width=\linewidth]{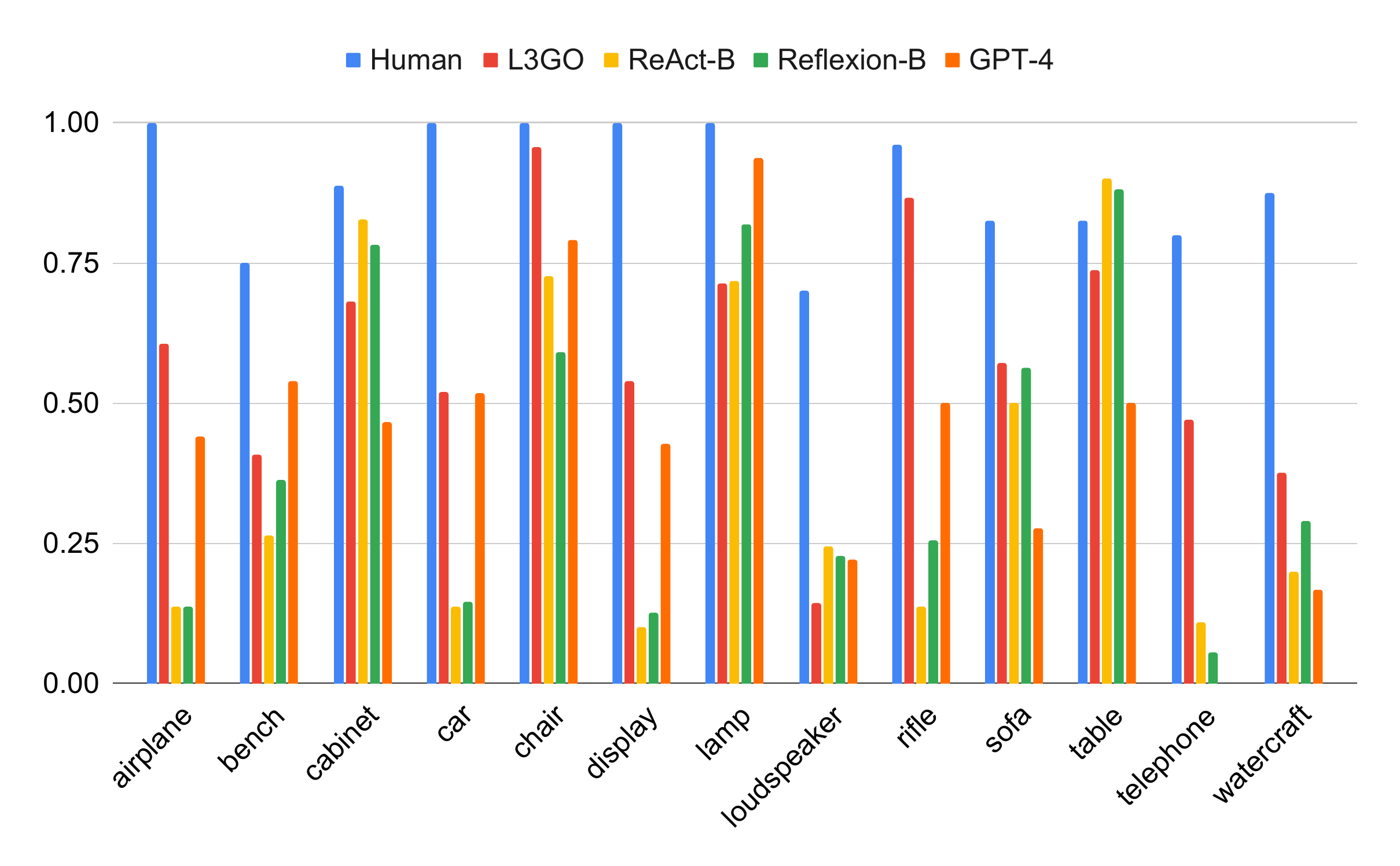}
    \vspace{-7mm}
    \caption{Human evaluation of \METHODNAME, ReAct-B, Reflexion-B, and GPT-4 on ShapeNet-13. `Human' refers to the original human-designed ShapeNet meshes. We observed a pattern similar to that in GPT-4V's evaluation.
    }
    \label{fig:human_eval_shapenet}
    \vspace{-4mm}
\end{figure}

We recruit human participants to assess whether the evaluation of GPT-4V aligns with human judgment.
For each mesh, we use 10 images from 10 different angles (the same images as above for GPT-4V evaluation), and ask a human participant to classify these images into one the 13 object categories. 
We collect 130 human responses by showing meshes generated by GPT-4, and \METHODNAME, as well as the original ShapeNet meshes, totaling in 390 human responses.
The category-by-category results are shown in Figure \ref{fig:human_eval_shapenet}.
We can see an overall pattern, where the original ShapeNet has high accuracy, \METHODNAME outperforms GPT-4, except for a few cases like ``lamp'', ``bench'', and ``loudspeaker''.

We also gather 390 responses from GPT-4V using the same set of images.
To benchmark GPT-4V against other top vision-language models, we also obtain 390 responses each from BLIP-2 and InstructBLIP.
Regarding human-to-human correlation evaluation, four participants were asked to classify all 390 images.
However, we exclude the data from the participant whose responses most often differed from the other three. 
We then calculate the Cohen's Kappa score to measure the correlation between these models and three human evaluators, averaging these correlations, as shown in Table \ref{tab:correlation_analysis}. 
Our findings indicate that GPT-4V's responses align most closely with human judgments, though it is important to note that even among humans, agreement was not perfect.

\begin{table}
  \centering
  \resizebox{\linewidth}{!}{
  \begin{tabular}{@{}ccccc@{}}
    \toprule
       & GPT-4V & InstructBLIP & BLIP-2 & Human \\
    \midrule
Human & $\textbf{0.512}_{(0.028)}$ & $0.344_{(0.016)}$ & $0.341_{(0.012)}$ & $0.569_{(0.020)}$ \\
    \bottomrule
  \end{tabular}
  }
  \caption{The Cohen's Kappa correlation between evaluations based on models and human judgement. We report the average and standard deviation calculated from three independent human evaluators.}
  \label{tab:correlation_analysis}
\end{table}

%% file: sec/6_experiment_unusual.tex
\section{Experiments on \TASKNAME: Constructing Unconventionally Feasible Objects}
\label{sec:exp_owva}

Our previous experiments show that \METHODNAME can accurately construct a simple object from ShapeNet around 60\% of the time. However, modern diffusion based models can nearly always generate an image of a given ShapeNet object. This is in part because there are many possible valid instantiations of, e.g., ``car" or ``bench". So: is there any potential practical advantage to using a method like \METHODNAME?

To illustrate the potential advantages of LLMs in spatial construction, we introduce a benchmark that requires more precise spatial understanding. %
Inspired by DrawBench \cite{saharia2022} and PartiPrompts \cite{yu2022tmlr} which are collections of prompts that systematically evaluate text-to-image models, we introduce \TASKNAME: a set of 50 difficult prompts that 1) require precise spatial understanding to construct; and 2) are unusual, in the sense that they are less likely to occur during text-only pre-training, e.g., ``a chair with one armrest''. %

The prompts in \TASKNAME span 9 object categories, and each prompt is a combination of a common object with varied characteristics such as ``a chair with five legs'', ``a mug with two handles'' and so on.
The full prompt list is shown in the Appendix.
We focus on everyday 3D objects to help us isolate the model's performance in accurately interpreting prompts from its inherent ability to create unconventional objects. 
By using simple-to-assemble items such as sofas, chairs, lamps, and tables, we can better discern whether any shortcomings are due to the model's prompt following from its object creation skills.

\paragraph{Baselines}
We compare our LLM-based approach with latest text-to-2D  and text-to-3D methods such as DALL-E 3 \cite{betkerh}, Stable Diffusion XL (SDXL) \cite{podell2023}, and Shap-E \cite{jun2023}. 
DALL-E 3 uses descriptive synthetic captions to improve prompt following of DALL-E 2 \cite{ramesh2022}, the previous version of OpenAI's text-to-image diffusion model. 
Stable Diffusion XL is an open-sourced text-to-image diffusion model. 
Shap-E \cite{jun2023} is a text-to-3D model that generates the parameters of implicit function for 3D models which then can be rendered.
Since DALL-E 3 automatically re-writes the prompt for safety reasons and adds more detail, (and it is not possible to disable this feature at the moment) we add ``I NEED to test how the tool works with extremely simple prompts. DO NOT add any detail, just use it AS-IS:'' to our prompt as recommended by OpenAI \footnote{https://platform.openai.com/docs/guides/images}. %

\paragraph{Experiment procedures}

We again utilize the judgements of human participants to evaluate the output of our models on \TASKNAME.
For each given prompt, we generate 10 random objects from one model and another 10 from a different model.
A participant is then asked to judge which set of images better matches the provided text caption. 
If they believe that neither set accurately represents the caption, they can choose ``no preference.''
For each experiment, we recruit 10 human evaluators.
Additionally, we include 4 attention check questions on top of the 50 total questions. 
Any evaluator who does not correctly answer all the attention check questions is excluded from our analysis.
For models that create 3D objects, we render 10 images from various angles by rotating the camera at a constant height. 
These 10 images are then compiled into a rotating GIF.

\paragraph{Results}

\begin{figure}[!t]
    \centering
    \includegraphics[width=\linewidth]{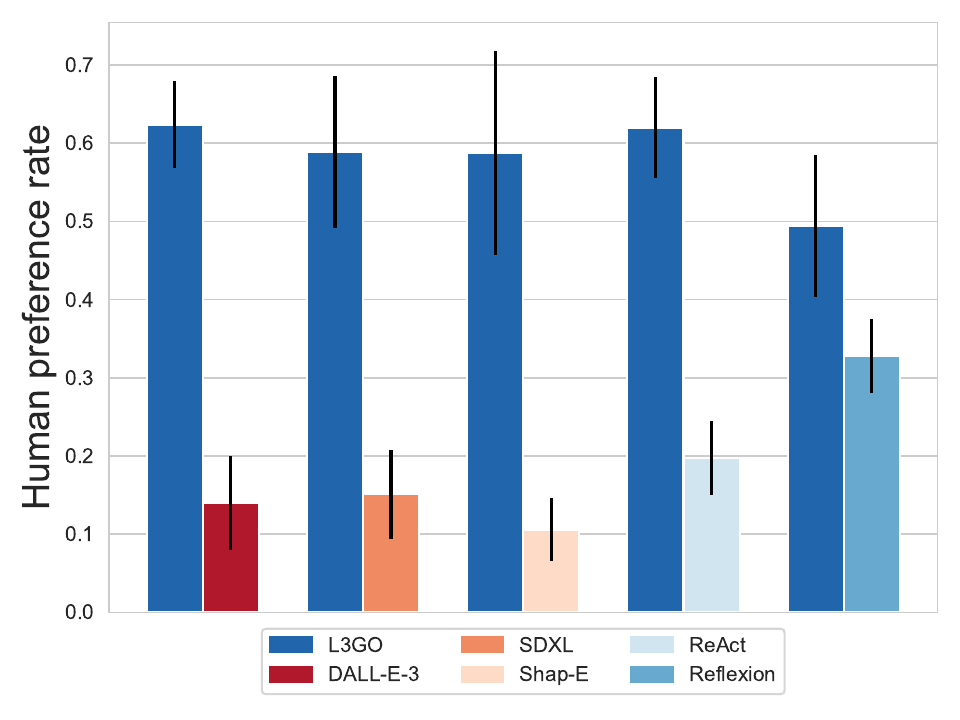}
    \vspace{-7mm}
    \caption{Human preference of \METHODNAME vs. DALL-E-3, SDXL, Shap-E, ReAct-B, and Reflexion-B on \TASKNAME.}
    \label{fig:human_pref_unusual}
    \vspace{-4mm}
\end{figure}

\begin{figure*}[!t]
    \centering
    \includegraphics[width=\linewidth]{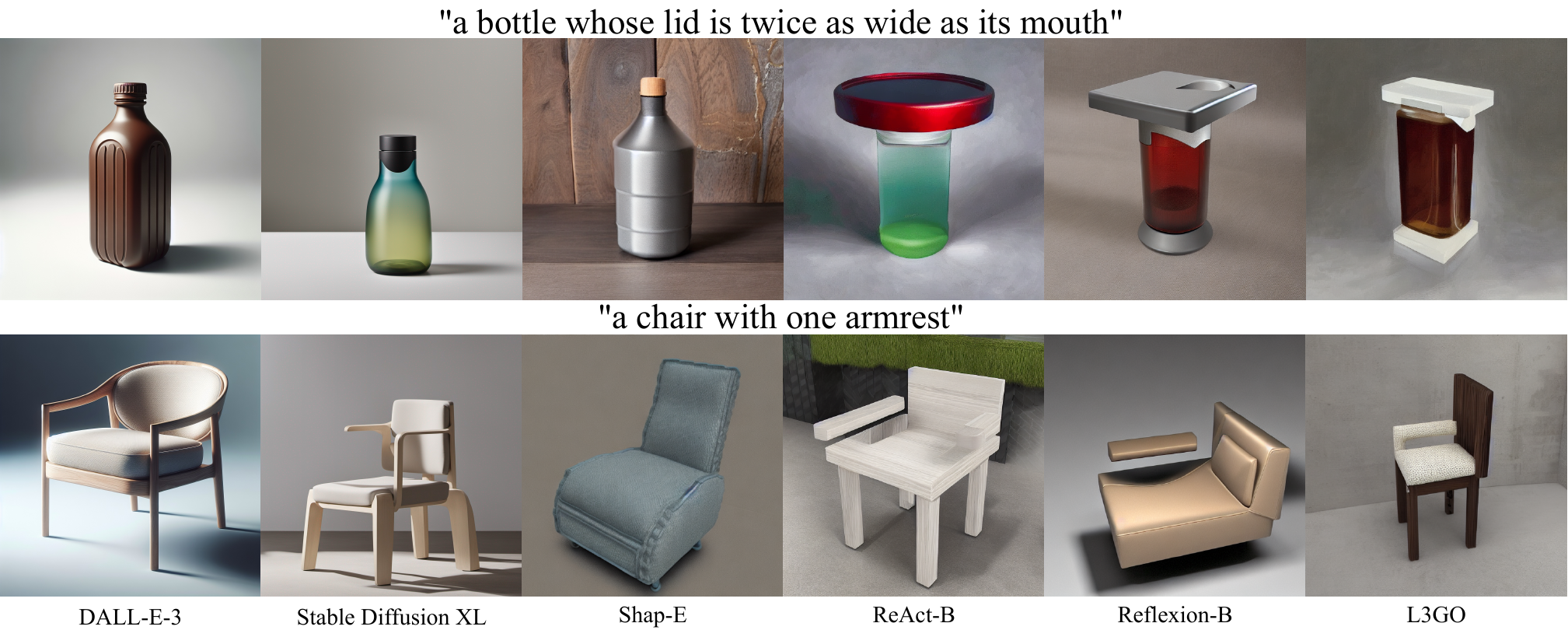}
    \vspace{-7mm}
    \caption{Example generated images based on \TASKNAME. The LLM-based approaches (ReAct-B, Reflexion-B, and \METHODNAME) successfully create the desired objects, while some of the most advanced text-to-image and text-to-3D models (DALL-E 3, Stable Diffusion XL, and Shap-E) still struggle to follow the prompt perfectly. %
    }
    \label{fig:ufo_examples}
    \vspace{-4mm}
\end{figure*}

The results are shown in Figure \ref{fig:human_pref_unusual}.
\METHODNAME outperforms the other LLM agents (e.g. ReAct-B, Reflexion-B) and the state-of-the-art text-to-image models (DALL-E-3 and SDXL) and text-to-3D model (Shap-E) in terms of human preference.
Example generated images are shown in Figure \ref{fig:ufo_examples}.
DALL-E-3, SDXL and Shap-E produce images that do not perfectly follow the specific prompt instructions. 
While SDXL is able to generate a desk with three legs, an additional chair that is not asked in the prompt is generated.
DALL-E-3 seems to completely ignore the specific requirements of prompts.
In contrast, while their designs are not perfect, language model-based agents are capable of constructing chairs with the right number of legs. 
These results suggest that structured reasoning may serve as a viable strategy to mitigate the challenges posed by insufficient training data.

\paragraph{Effect of background and texture on evaluation for \TASKNAME}

\begin{figure}[!t]
    \centering
    \includegraphics[width=\linewidth]{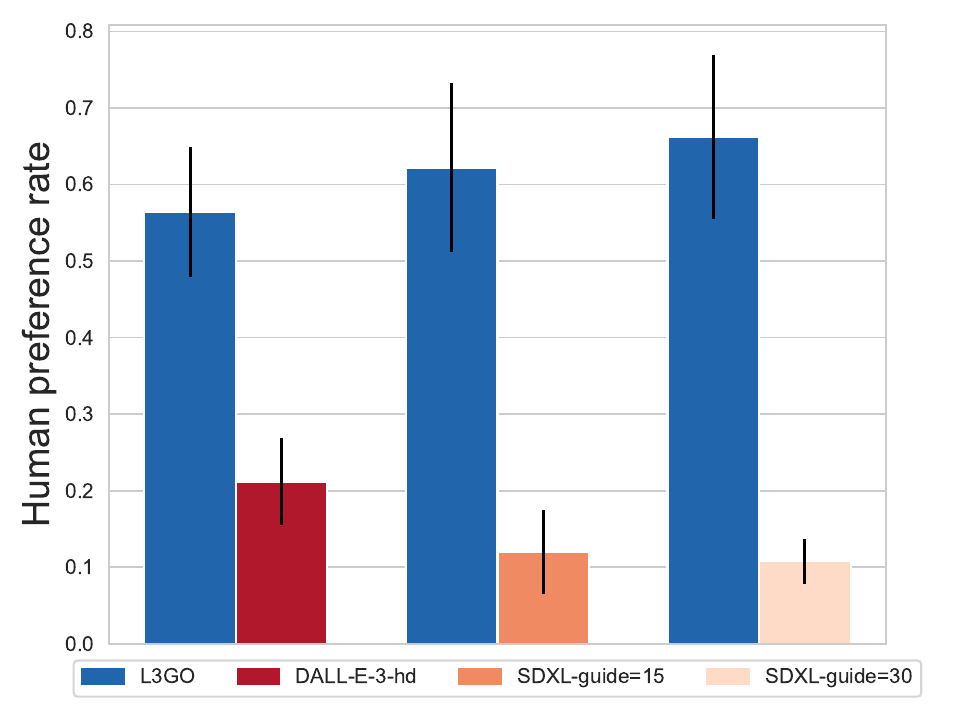}
    \vspace{-7mm}
    \caption{Human preference of \METHODNAME vs. DALL-E-3 and SDXL on \TASKNAME, where we attempt to make the background of generated images to be simple gray shading. We also vary guidance scales for SDXL to see if better prompt following improves the performance, denoted as `guide=15, 30'. For DALL-E-3, we use `quality=hd' option for enhanced detail.}
    \label{fig:human_pref_unusual_black_bg}
    \vspace{-4mm}
\end{figure}

We look into how the background and texture differences in images created by text-to-image models and LLM-based methods might affect human evaluations. 
To test this, we change prompt styles with these text-to-image models.
For DALL-E-3, we use ``[object name] Make sure the background is black, and the object is a gray-colored 3D shape.'' 
For Stable Diffusion XL, we use ``[object name], black background, gray-colored 3D shape.'' 
Additionally, we alter the guidance scale for Stable Diffusion XL, which determines how closely the diffusion model follows the text prompts. 
In both scenarios, we observe that \METHODNAME outperforms text-to-image models in terms of human preference, as illustrated in Figure \ref{fig:human_pref_unusual_black_bg}.
We also conducted an initial test to determine if GPT-4V could serve as an evaluator for \TASKNAME. 
However, we observed that in over 20\% of the cases, GPT-4V refuses to provide an answer, which might be related to the characteristics of the images generated.
We refer to Table \ref{tab:gpt_4v_eval_unusualbench} in Appendix for more details.

%% file: sec/8_ablation.tex
\section{Ablation studies}
\label{sec:ablation}

We ablate three 3 system design choices to see which component most impacts the overall performance for 3D mesh generation. 
We use our automatic evaluation via GPT-4V to compare the performance in 3 ablations: 
1) without spatial critic, 
2) without program-based coordinate calculator, and
3) choice of LLMs.
For each setting, we generate 10 objects per category, and render 10 images from different angles. 
For 1) and 2), the mean accuracy across the 13 categories of ShapeNet (evaluated by GPT-4V) are 0.515 and 0.585, respectively.
In comparison, \METHODNAME achieves a higher score of 0.6.
While the average scores for \METHODNAME and without coordinate calculator achieve similar scores, the latter scores 0 in both the cabinet and car categories.
In contrast, when \METHODNAME employs program-based coordinate calculation, it achieves scores of 0.5 and 0.4 for these categories, respectively.

\begin{table}
  \centering
  \resizebox{\linewidth}{!}{
  \begin{tabular}{@{}ccccc@{}}
    \toprule
       & Mixtral-8x7B & w/o spatial critic & w/o program-based & \METHODNAME \\
    \midrule
 & 0.138 & 0.515 & 0.585 & \textbf{0.6} \\
    \bottomrule
  \end{tabular}
  }
  \vspace{-4mm}
  \caption{Ablation studies. We evaluate the performance based on ShapeNet's 13 categories using GPT-4V as an evaluator; each cell is an average over 130 trials. `w/o spatial critic/program-based' refers to \METHODNAME based on GPT-4 without spatial critic and without program-based coordinate calculation module. `Mixtral-8x7B' refers to ReAct-B based on Mixtral-8X7B instead of GPT-4.
  }
  \label{tab:correlation_analysis}
  \vspace{-6mm}
\end{table}

\paragraph{Open-sourced LLMs}
We explore the use of open-source LLMs in place of GPT-4. 
For this purpose, we use Mixtral-8x7B, a sparse mixture-of-experts model \cite{jiang2024}, that is known to either match or surpass the performance of Llama-2 70B and GPT-3.5 in most benchmark tests.
We carry out experiments using ReAct-B and ShapeNet-13, with GPT-4V serving as our evaluation tool. 
While the accuracy for the ReAct-B(Mixtral-8x7B) for most shape categories ranged between 0 and 0.1, the categories of sofas, lamps, and tables achieved higher scores of 0.3, 0.3, and 0.7, respectively.
This is likely due to their simpler shapes and easier recognizability.
The average accuracy score is 0.138.
This result is significantly lower than the 0.423 accuracy achieved by ReAct-B(GPT-4), as indicated in Table \ref{tab:gpt-4v_eval_shapenet-13}.
This indicates that the task of constructing mesh objects, which demands a precise understanding of 3D space, still needs the reasoning abilities found at the level of GPT-4.

%% file: sec/11_conclusion.tex
\section{Conclusion}
\label{sec:conclusion}

We introduced \METHODNAME, a language agent designed to generate 3D objects from text instructions through an API we developed for Blender, a 3D modeling software. Our evaluation using 13 largest object categories from ShapeNet shows that \METHODNAME's superior capabilities in comparison to other models such as GPT-4, ReAct, and Relfexion. Additionally, we devised \TASKNAME, a set of challenges aimed  at testing the ability of generative AI models in creating common objects with unconventional characteristics. 
The performance of \METHODNAME marks a significant advancement in the application range of language models. 
For instance, diffusion models could be further improved with unconventional data generated by structured prompting. 
Moreover, analyzing how language models process spatial information with internal model representations may yield valuable insights into understanding and improving their 3D modeling abilities.

%% file: sec/impact_statement.tex
\paragraph{Impact Statement}

Our research indicates the vast potential for integrating language models with 3D modeling, potentially revolutionizing design processes and the creation of digital environments. This convergence aims at making generative AI tools more intuitive and capable of supporting creative endeavors. With \METHODNAME, we begin to tap into the untapped possibilities in this domain, setting the stage for extensive future exploration and development.

The positive societal impacts of our work could be substantial, particularly in design, engineering, and the arts, by enabling the visualization and prototyping of ideas that were previously difficult or impossible to achieve. Furthermore, our approach could enhance educational tools, making complex concepts more accessible through interactive and visually intuitive representations.
However, we must also consider the ethical implications of advancing image generation technology, such as the potential for creating misleading or harmful content. It underscores the necessity for ongoing research into mechanisms that ensure these powerful tools are used responsibly and ethically. We advocate for a balanced approach that emphasizes innovation alongside responsibility and ethical considerations.

%% file: sec/12_appendix.tex
\section{Algorithm}

\begin{figure}[!t]
    \centering
    \includegraphics[width=0.7\linewidth]{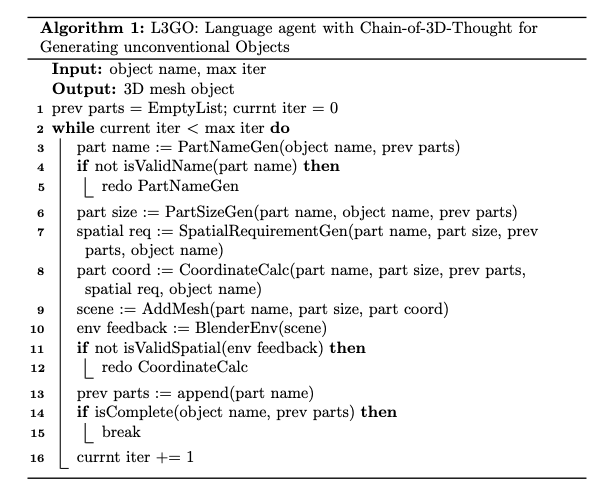}
    \caption{Pseudo algorithm of \METHODNAME.}
    \label{fig:algorithm_our_approach}
\end{figure}

The pseudo algorithm of \METHODNAME is given in Figure \ref{fig:algorithm_our_approach}.

\section{GPT-4V as an evaluator for \TASKNAME}

We also conducted an initial test where we used GPT-4V to evaluate generated images using prompts from \TASKNAME. 
This experiment involved showing GPT-4V two sets of 10 images from two models, similar to our human study, and asking GPT-4V which set of images more accurately represent the corresponding text caption.
We ask it choose either the 'top row', 'bottom row', or 'no preference', along with providing a reason for its choice. 
The images were generated using \METHODNAME and DALL-E-3. 
We carried out this experiment twice. 
In both cases, for more than 20\% of the prompts (13 and 18 out of 50), the response we received was ``I cannot assist with this request''. 
Due to this, we decided to rely on human evaluators instead of GPT-4V for our main experiments.

\begin{table}
  \centering
  \resizebox{0.6\linewidth}{!}{
  \begin{tabular}{@{}cccccc@{}}
    \toprule
       & \METHODNAME & DALL-E-3 & No pref  & Refuse &  No images \\
    \midrule
1st trial & 12 & 8 & 9 & 18 & 3  \\
2nd trial & 14 & 12 & 11 & 13 & 0 \\
    \bottomrule
  \end{tabular}}
  \caption{GPT-4V's evaluation of \TASKNAME, where GPT-4V compares the image generation by \METHODNAME and DALL-E-3. ``Refuse'' indicates instances where GPT-4V declined to respond, for example by saying ``I cannot assist with this request.'' ``No images'' refers to cases where GPT-4V incorrectly perceived the images as being invisible.}
  \label{tab:gpt_4v_eval_unusualbench}
\end{table}

\section{Details for human evaluation}

We recruit participants through Prolific.com, an online platform used by researchers to conduct behavioral studies and gather participants. The requirement for participants was English fluency, as the study's materials were all in English. The study was made accessible to all qualified participants on the platform.
We ensure \$15 hourly wage.

At the beginning of the experiment, we give participants an online form detailing the experiment's aims, a summary, the methodology, confidentiality assurances, and the voluntary nature of their involvement. The participants are assured of the confidentiality of their responses and their right to withdraw at any time. Participants are asked to acknowledge their understanding of the provided information and their consent to participate by clicking the ``start'' button. The experiment commences only after this button was clicked.

Despite efforts to minimize bias by keeping participant criteria to a minimum, there remains an inherent bias towards English-speaking individuals with access to computers and the internet. We note that this bias also tends to reflect characteristics of the WEIRD (Western, Educated, Industrial, Rich, Democracies) demographic.

\section{Additional details for experiments}

\begin{figure}[!t]
    \centering
    \includegraphics[width=0.7\linewidth]{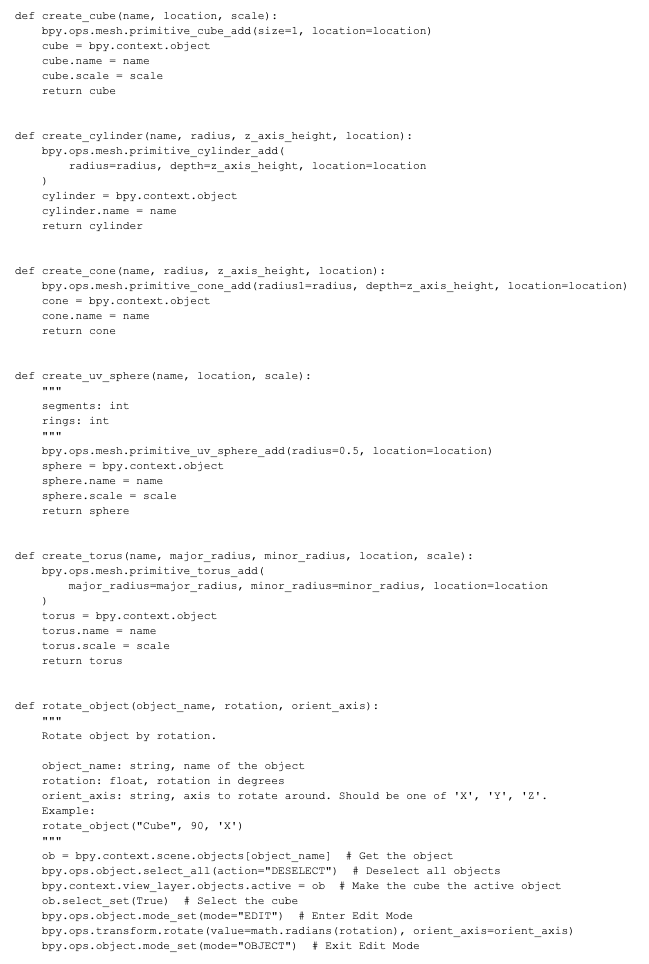}
    \caption{Full list of action wrapper APIs for \ENVNAME.}
    \label{fig:list_action_apis}
\end{figure}

\paragraph{Wrapper APIs for action commands in \ENVNAME} are shown in Figure \ref{fig:list_action_apis}.

\begin{figure}[!t]
    \centering
    \includegraphics[width=0.7\linewidth]{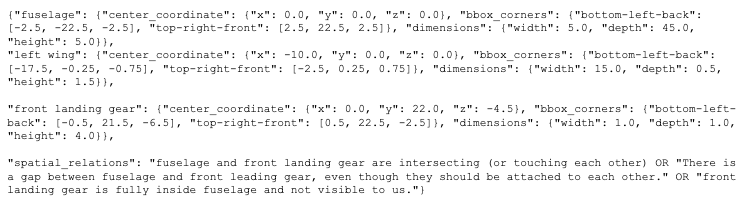}
    \caption{An example environment feedback from \ENVNAME.}
    \label{fig:example_env_feedback}
\end{figure}

\paragraph{Example environment feedback from \ENVNAME} is shown in Figure \ref{fig:example_env_feedback}.

\begin{figure}[!t]
    \centering
    \includegraphics[width=0.7\linewidth]{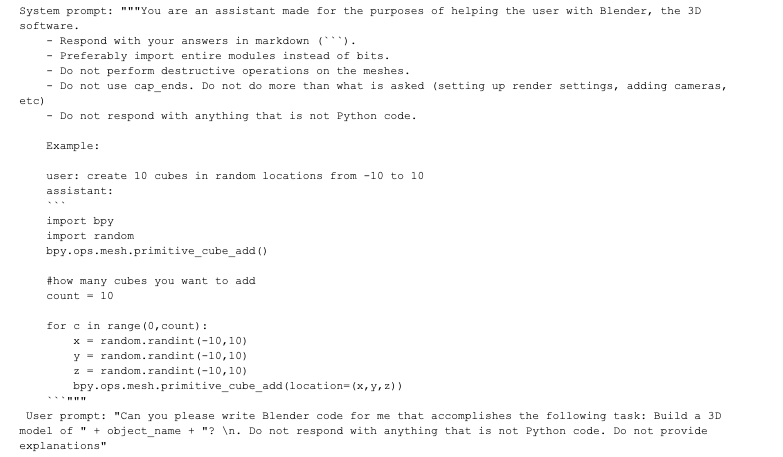}
    \caption{The full prompt for the baseline GPT-4 used in our experiments. This prompt is taken from Blender-GPT\footnote{\url{https://github.com/gd3kr/BlenderGPT}}, which is the first open-source code that uses GPT-4 inside Blender to our knowledge.}
    \label{fig:prompt_for_gpt4}
\end{figure}

\paragraph{Full prompt for the baseline GPT-4} is shown in Figure \ref{fig:prompt_for_gpt4}.

\paragraph{The list of prompts in \TASKNAME}
a chair with two legs,
a chair with three legs,
a chair with five legs,
a chair with six legs,
a chair with seven legs,
a chair with one armrest,
a chair with three armrests,
a chair with two backrests,
a chair with three backrests,
a stool with two legs,
a stool with three legs,
a stool with five legs,
a stool with six legs,
a stool with seven legs,
a stool with one armrest,
a stool with three armrests,
a stool with two backrests,
a stool with three backrests,
a desk with two legs,
a desk with three legs,
a desk with five legs,
a desk with six legs,
a desk with seven legs,
a table with two legs,
a table with three legs,
a table with five legs,
a table with six legs,
a table with seven legs,
a pair of eyeglasses with one round lens and one square lens,
a pair of eyeglasses with one round lens and one triangle lens,
a pair of eyeglasses with one square lens and one triangle lens,
a pair of eyeglasses with three lenses,
a pair of eyeglasses with four lenses,
a pair of eyeglasses with five lenses,
a sofa with one leg,
a sofa with two legs,
a sofa with three legs,
a sofa with five legs,
a sofa with legs that are longer than its backrest,
a sofa with armrests that are longer than its backrest,
a lamp with two legs,
a lamp with four legs,
a lamp with five legs,
a bottle whose lid is twice as wide as its mouth,
a bottle with a lid that is three times wider than its mouth,
a bottle with a lid that is four times wider than its mouth,
a mug with two handles,
a mug with three handles,
a mug with four handles,
a mug with five handles

\subsection{Additional generated examples} 
\begin{figure}[!t]
    \centering
    \includegraphics[width=\linewidth]{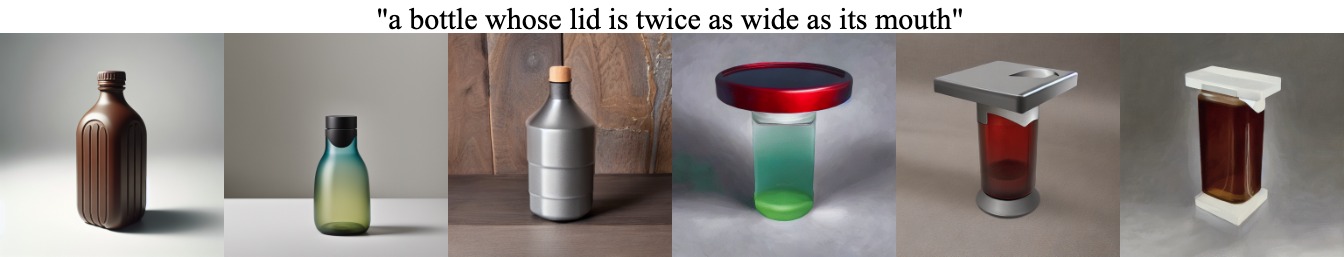}
    \caption{Additional generated examples on \TASKNAME.}
    \label{fig:more_examples}
\end{figure}

\begin{figure}[!t]
    \centering
    \includegraphics[width=\linewidth]{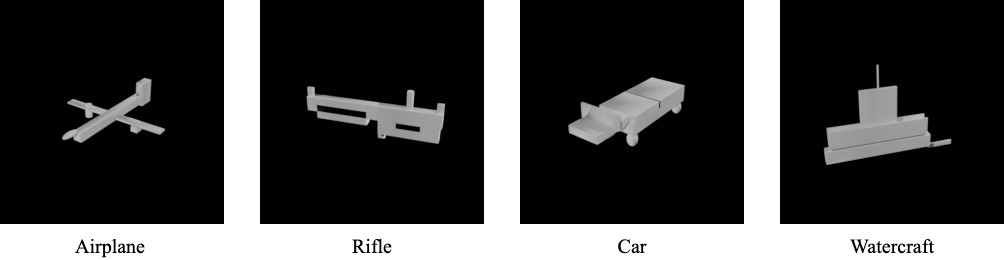}
    \caption{Additional generated examples of \METHODNAME on ShapeNet.}
    \label{fig:shapenet_lego_examples}
\end{figure}

are shown in Figure \ref{fig:more_examples} and \ref{fig:shapenet_lego_examples}.

The code is uploaded in the following url.\footnote{\url{https://u.pcloud.link/publink/show?code=kZ4e550Z2VnyJfJO8VHyivUkfVdKhb5B8SEk}}

%% file: sec/9_app_future.tex
\section{Applications and Future Work}
\label{sec:applications_future_work}

\subsection{Creative Combinations of Object Parts}
\label{sec:creative}

In \TASKNAME, we explored how our approach could adapt to unique variations in common object characteristics, such as creating a chair with an atypical number of legs. 
While legs are a standard feature of chairs, our focus was on experimenting with unconventional leg counts. 
In this section, we extended this question to see if \METHODNAME can handle even more creative combinations, like integrating components not typically associated with the object, for instance, a chair with wings.

\begin{figure}[!t]
    \centering
    \includegraphics[width=\linewidth]{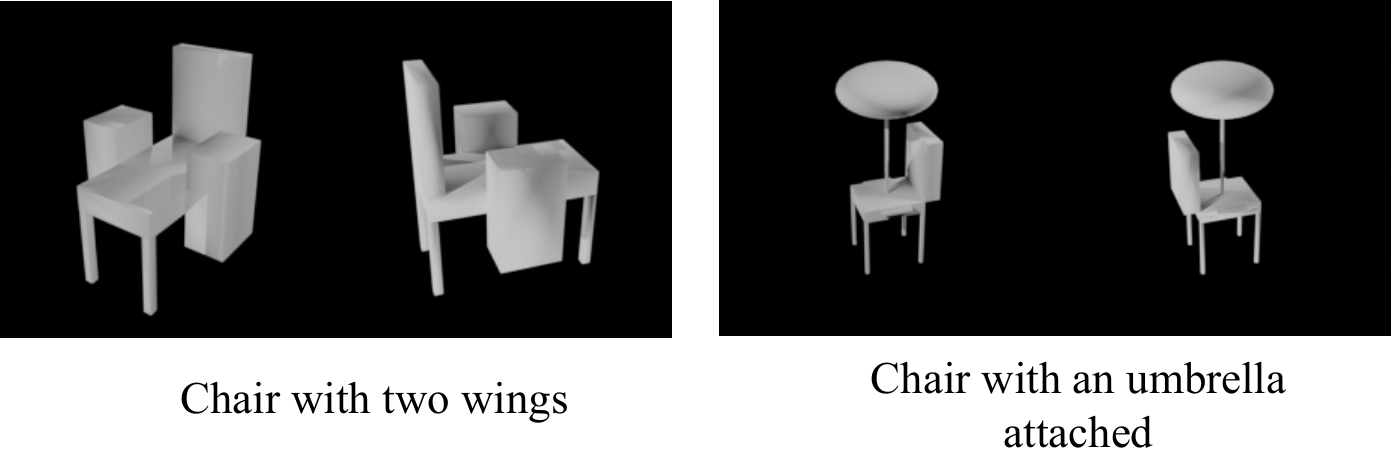}
    \caption{Because Large Language Models can interpret creative prompts, \METHODNAME can create unique items such as a chair with wings or a chair that has an umbrella attached.}
    \label{fig:creative_examples}
\end{figure}

Displayed in Figure \ref{fig:creative_examples} are instances of these creative prompts and their resulting models. 
Even though the forms for ``wings'' and ``umbrella'' are basic, such as rectangles and elongated spheres with thin cylinders, \METHODNAME adeptly figures out that ``wings'' should attach to the sides of the chair, and an ``umbrella'' should be positioned on top.
(See Appendix for more examples.) 
Our results demonstrate that \METHODNAME can successfully interpret and construct these unusual designs.

\subsection{Language interface for complex softwares}

\begin{figure}[!t]
    \centering
    \includegraphics[width=\linewidth]{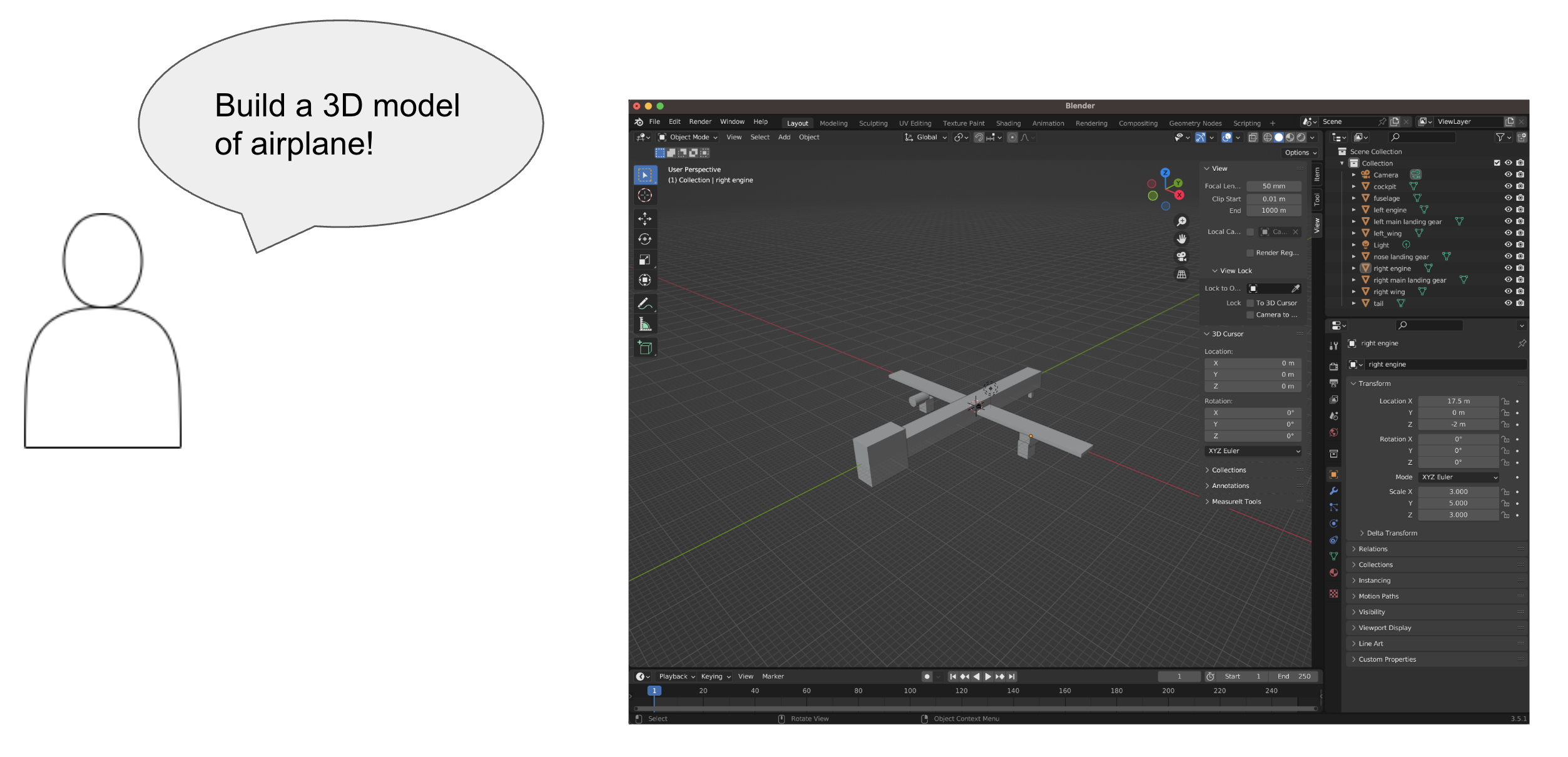}
    \caption{The Blender user interface can be daunting for newcomers. An LLM agent can be helpful by creating a draft of an object to assist beginners.}
    \label{fig:lang_interface_example}
\end{figure}

Most 3D modeling software programs have a steep learning curve. 
Creating sophisticated 3D models often involves a sequence of operations using complex interfaces, which can be challenging for non-experts to navigate. 
Our method can be used as a language interface for these complex software programs, making them more accessible to beginners. 
As illustrated in Figure \ref{fig:lang_interface_example}, a user can input a prompt, and \METHODNAME will generate an initial draft of the object. 
While our approach is still in its early stages and needs further development for practical use, we hope our research inspires future advancements.

%% file: sec/10_limitation.tex
\section{Limitations}
\label{sec:limitations}

The quality of 3D mesh objects generated using LLM-based methods lag behind those produced by diffusion-based text-to-3D methods, and far from being on par with human standards.
Moreover, creating simple objects takes roughly 3 to 5 minutes, while more complex ones can take 10 to 20 minutes, depending on the number of retry operations needed after the feedback.
We believe that the future work should explore more efficient LLM-based approaches to create 3D mesh as this is a promising direction to better control the eventual 3D object generation.